\documentclass{article}

    \PassOptionsToPackage{numbers, compress}{natbib}

\usepackage[preprint]{neurips_2025}

\usepackage[utf8]{inputenc} 
\usepackage[T1]{fontenc}    
\usepackage{hyperref}       
\usepackage{url}            
\usepackage{booktabs}       
\usepackage{amsfonts}       
\usepackage{nicefrac}       
\usepackage{microtype}      
\usepackage{xcolor}         
\usepackage{multirow} 
\usepackage{graphicx}
\usepackage{wrapfig}
\usepackage{amsmath} 
\usepackage{tikzlings-marmots}

\title{MARMOT: Masked Autoencoder for Modeling Transient Imaging}

\author{%
  Siyuan~Shen\\
  ShanghaiTech University\\
  \texttt{shensy2023@shanghaitech.edu.cn} \\
  \And
  Ziheng~Wang \\
  ShanghaiTech University\\
  \texttt{wangzh1@shanghaitech.edu.cn} \\
  \And
  Xingyue~Peng \\
  ShanghaiTech University\\
  \texttt{pengxy2023@shanghaitech.edu.cn} \\
  \And
  Suan~Xia \\
  ShanghaiTech University\\
  \texttt{xiasa2022@shanghaitech.edu.cn} \\
  \And
  Ruiqian~Li \\
  ShanghaiTech University\\
  \texttt{lirq1@shanghaitech.edu.cn} \\
   \And
  Shiying~Li \\
  ShanghaiTech University\\
  \texttt{lishy1@shanghaitech.edu.cn} \\ 
\And
  Jingyi~Yu \\
  ShanghaiTech University\\
  \texttt{yujingyi@shanghaitech.edu.cn} \\
}

\begin{document}

\maketitle

\begin{abstract}
Pretrained models have demonstrated impressive success in many modalities such as language and vision. Recent works facilitate the pretraining paradigm in imaging research. Transients are a novel modality, which are captured for an object as photon counts versus arrival times using a precisely time-resolved sensor. In particular for non-line-of-sight (NLOS) scenarios, transients of hidden objects are measured beyond the sensor's direct line of sight. Using NLOS transients, the majority of previous works optimize volume density or surfaces to reconstruct the hidden objects and do not transfer priors learned from datasets. In this work, we present a masked autoencoder for modeling transient imaging, or MARMOT, to facilitate NLOS applications. Our MARMOT is a self-supervised model pretrianed on massive and diverse NLOS transient datasets. Using a Transformer-based encoder-decoder, MARMOT learns features from partially masked transients via a scanning pattern mask (SPM), where the unmasked subset is functionally equivalent to arbitrary sampling, and predicts full measurements. Pretrained on TransVerse—a synthesized transient dataset of 500K 3D models—MARMOT adapts to downstream imaging tasks using direct feature transfer or decoder finetuning. Comprehensive experiments are carried out in comparisons with state-of-the-art methods. Quantitative and qualitative results demonstrate the efficiency of our MARMOT.  
\end{abstract}
\newcommand{\shensy}[1]{{\color{magenta}{[Siyuan: #1]}}}
\newcommand{\cn}[1]{{\color[RGB]{0,0,0}{\begin{CJK}{UTF8}{gbsn} #1 \end{CJK}}}}

\newcommand{\Li}[1]{{\color{blue}{[Li: #1]}}}

\newcommand{\transients}{\tau}

\maketitle


\section{Introduction}
\label{sec:intro}

Pretrained models have achieved unprecedented progress in vision and beyond. Being pretrained on massive and diverse datasets, these models capture intricate data patterns and relationships for strong generalization capabilities. The learned features are applicable and adaptive in a wide range of downstream tasks. Prominent examples include Segment Anything Model~\cite{kirillov2023segment} for image segmentation and Diffusion Model~\cite{ho2020denoising,song2020denoising} for image generation. Inspired by success in many data modalities, e.g., language, image, and video, the pretraining paradigm becomes the driving force to advance imaging research. Pretrained models in imaging modalities, including ultrasound~\cite{jiao2023usfm} and radiographic lesions~\cite{pai2024foundation, ma2024segment}, facilitate efficient learning for tasks such as lesion segmentation and disease diagnosis. In fluorescence microscopy imaging modality, Ma et al.~\cite{ma2024pretraining}
show the effectiveness of their pretrained model via finetuning on five image restoration tasks.

\begin{figure}[t]
\begin{center}
\includegraphics[width=1.0\linewidth]{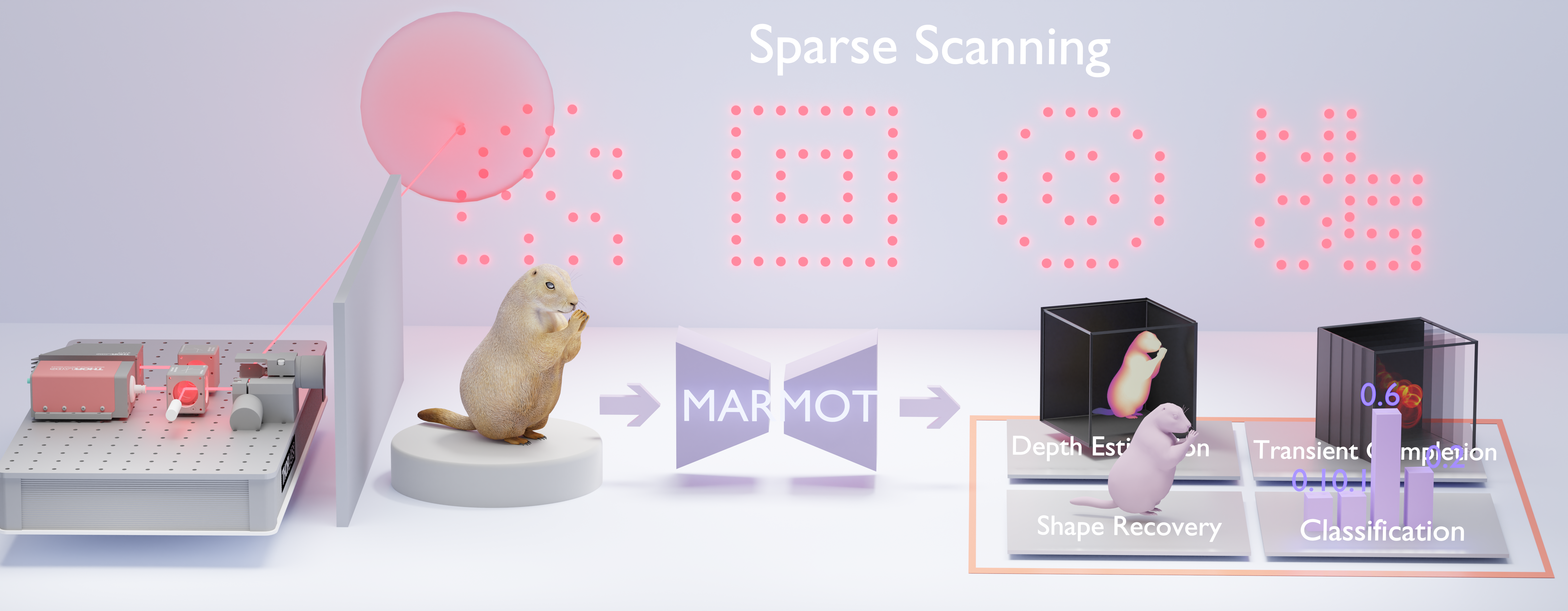}
\end{center}
  \caption{\textbf{Concept of MARMOT.} We resort to a masked autoencoder for modeling transients. In particular in non-line-of-sight (NLOS) scenarios, properties of hidden objects are measured through spherical light trajectories. MARMOT encodes sparse, non-uniform transients (e.g., red scanning points on the wall) into latent features and reconstructs the full transient field via a decoder. Using the features and the recovered transients, MARMOT adapts to a wide range of NLOS tasks.
}
\label{fig:teaser}
\end{figure}

\begin{figure*}[t]
\begin{center}
\includegraphics[width=1.0\linewidth]{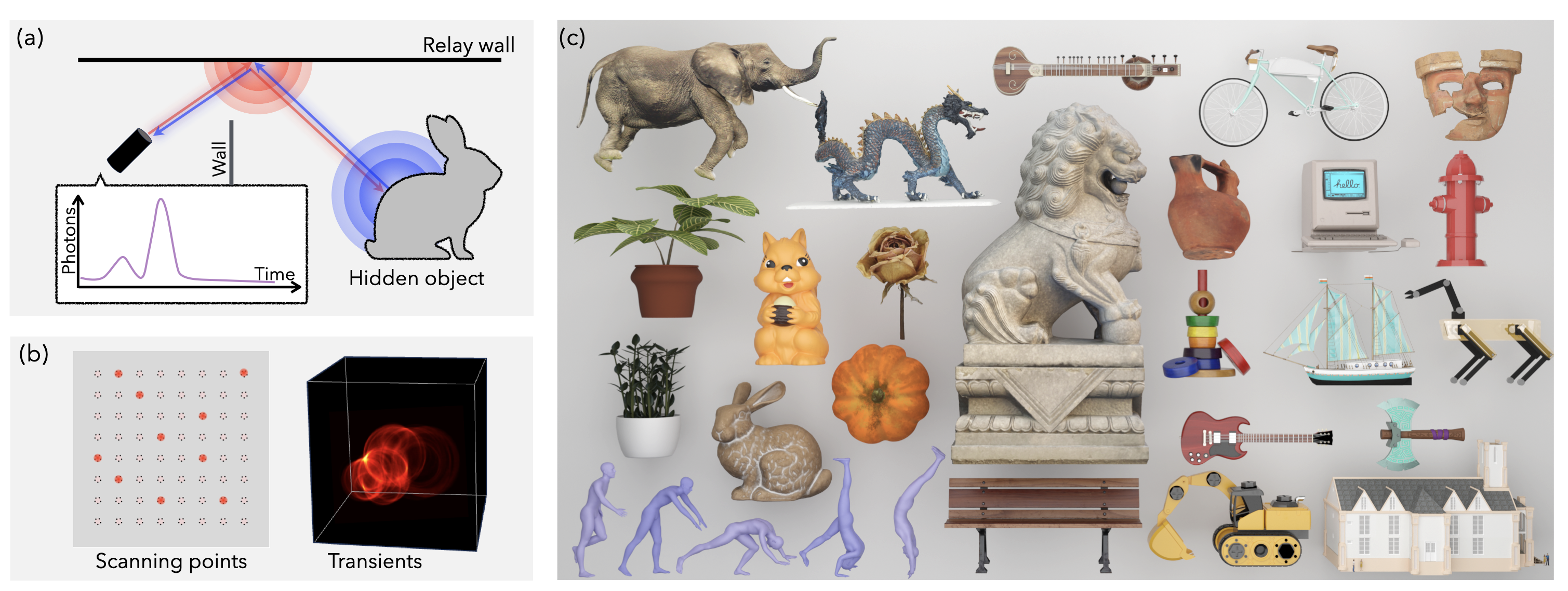}
\end{center}
  \caption{\textbf{(a) Schematic of confocal NLOS imaging.} A laser is directed towards a relay wall where light propagates in spherical wavefronts. After hitting the object, light bounces off back toward the wall. A time-resolved detector records transients, which indicates photon counts over time bins.
\textbf{(b) Scanning points and transients} By scanning the points on the wall, transients are collected. 
\textbf{(c) Exemplars of TransVerse dataset.} We select 500K 3D objects from Objaverse to generate our dataset, in which objects have varieties of properties, such as geometry and style.
}
\label{fig:datasets}
\end{figure*}

In this work, we explore the potential of pretrained models in transient, which is recorded using a time-resolved sensor at picosecond or nanosecond resolution. Conventional line-of-sight (LOS) imaging captures RGB images or transients of an object along a direct ray. 
Non-line-of-sight (NLOS) imaging, in contrast, indirectly measures transients of a hidden object behind obstacles or around corners~\cite{kirmani2009,velten2012NC}. Figs.~\ref{fig:teaser} and~\ref{fig:datasets} (left) illustrate a typical NLOS scenario. 
An ultrafast pulsed laser illuminates a relay surface point, generating spherical wavefronts that reach the hidden object and scatter back to relay surface. A single-photon sensor records a histogram (i.e., a transient) of photon counts over time bins. 
The measured transients contain rich information, which is convolved into spherical light trajectories. As a result, recovering hidden objects is a computationally prohibitive inverse problem. backprojection (BP)~\cite{velten2012NC} and its variants discretize the object into voxels and compute the posterior probability that transients are intersected with the voxels. Physics-based methods such as light-cone transform (LCT)~\cite{otoole2018LCT} and $f-k$ migration (f-k)~\cite{lindell2019fk} simplify the light path by considering the sensor and the laser collocate and resolve the reconstruction problem as 3D convolution. On the other hand, neural fields-based methods, e.g., neural transient fields (NeTF)~\cite{shen2021NeTF}, advance NLOS reconstruction, but they only exploit neural networks as an optimizer and do not facilitate priors learned from datasets, resulting in limited reconstruction quality due to the NLOS forward model.    

We resort to the pretraining masked autoencoder, termed as MARMOT, to reconstruct hidden objects in an NLOS scene from sparse measurements. Delvlin et al.~\cite{devlin2019Bert} propose the masked language model and He et al.~\cite{he2022MAE} apply the idea of masked autoencoder (MAE) in vision. MAE is conceptually simple yet efficient: By masking a portion of data, a self-supervised learner predicts the full data from the residuals and learns meaningful features. Inspired by MAE, we tailor MARMOT as an end-to-end network composed of Transformer-based encoder and decoder. We consider the sparse transients measured under a scanning pattern mask (SPM) to be a random proportion of the complete transients, which are densely measured in uniform grids. During training, MARMOT employs the unmasked transients as input. The encoder estimates latent features, from which the decoder recovers overall transients. The core of MARMOT lies twofold: First, transients are spatial-temporal signals, differing from language (1D) and image (2D). Training on the complete transients is expensive in computation cost. MARMOT allows the encoder to only process a small portion of unmasked transients. This highly reduces the computation and memory cost. Second, MARMOT randomly masks transients of the complete set. This masking scheme resembles arbitrary scanning on the relay surface and enables MARMOT to handle flexible scanning patterns, in regular or irregular and sparse or dense samplings.  

We adapt MARMOT to various NLOS tasks through lightweight finetuning. The pretrained encoder provides transferable latent features for albedo estimation, depth recovery, and classification. For classification, only a few layers are appended; for albedo and depth, we adopt task-specific decoders inspired by prior works. Optionally, the encoder can be frozen to reduce training cost across tasks.


We synthesize a large-scale transient dataset to pretrain MARMOT. We select 500K 3D models from the publicly available Objaverse~\cite{deitke2023objaverse}. Using a CUDA-based simulator~\cite{chen2020learned, yu2023enhancing}, we generate confocal transients. The objects in our dataset, TransVerse, have varieties of geometry, and size and diversities ranging from daily life to cartoon-styled. In total, TransVerse includes one million transients. Furthermore, we add noise analog to an NLOS imaging system~\cite{pan2022onsite}. Due to transient sparsity and redundancy, we observe in experiments that MARMOT works well by masking a high proportion, e.g., $95\%$ of $64 \times 64$, which facilitates the overall pretraining time and resources.

We conduct comprehensive experiments on our large-scale TransVerse and existing public synthetic and real NLOS datasets. We evaluate MARMOT on transient completion, comparing reconstructed and ground-truth full transients under regular-grid scanning. Using the recovered transients, we compare our results with state-of-the-art (SOTA) methods for NLOS tasks, such as albedo estimation. We further assess the pretrained encoder’s transferability on downstream tasks including classification, albedo and depth estimation. Both quantitative and qualitative results show that MARMOT outperforms SOTA approaches and generalizes well across diverse NLOS datasets and tasks.

\section{Related Work}

Non-line-of-sight imaging has achieved remarkable progress in recent decade. We briefly review the most relevant works on imaging setup, reconstruction algorithms, and pretrained models. 


NLOS imaging recovers hidden objects from indirect light reflections~\cite{velten2012NC,kirmani2009}. Typical systems use a pulsed laser and single-photon sensor to capture photons scattered via a relay surface. Among existing implementations, the single-photon avalanche diode (SPAD) is relatively affordable and compact. A single-pixel SPAD, driven by a pair of galvanometers, can scan the relay surface in a regular pattern~\cite{buttafava2015non,isogawa2020efficient} or in arbitrary forms~\cite{pan2022onsite,liu2023NC}. O'Toole et al.~\cite{otoole2018LCT} present a confocal setting, where illumination and detection points coincide. Similar to the previous works~\cite{otoole2018LCT,shen2021NeTF, liu2023NC,lindell2019fk}, we also consider single SPAD-based NLOS imaging systems under the confocal configuration. 

Recovering hidden object is an ill-posed inverse problem. Numerous efforts have been made to address it~\cite{liu2019analysis, choi2023self, liao2021fpga}. We roughly distinguish between two categories: without and with priors. Without priors, light-cone transform~\cite{otoole2018LCT, young2020non} and wave-based solutions~\cite{lindell2019fk,liu2019PF} reconstruct hidden objects but require dense and regular transients. 
Backprojection-based methods project each bin onto voxels~\cite{velten2012NC,buttafava2015non,arellano2017Fast}.
NeTF~\cite{shen2021NeTF} and NLOS-NeuS~\cite{fujimura2023nlosneus} exploit neural fields to represent a hidden scene. These optimization approaches, including~\cite{chopite2020cvpr,liu2021non}, recover albedo and geometry from transients measured in arbitrary distributions while remaining high spatial resolution. Recently, several works leverage sparse transients for recovery~\cite{ye2021compressed,liu2023fewshot}. 
With physical priors, Learned Feature Embeddings~\cite{chen2020learned}, Attention-Guided Kernel (AGK)~\cite{yu2023enhancing}, NLOST~\cite{li2023nlost}, and NLOS3D~\cite{mu2022NLOS3d}
introduce end-to-end networks to learn features from synthetic datasets but still require dense measurements. More recent approaches employ sparser measurements~\cite{wang2023non,liu2023NC,li2023deep}, but they train on a size-limited dataset. Our MARMOT masks random scanning points, akin to an arbitrary sampling pattern. 

\begin{figure*}[t]
\begin{center}
\includegraphics[width=1.0\linewidth]{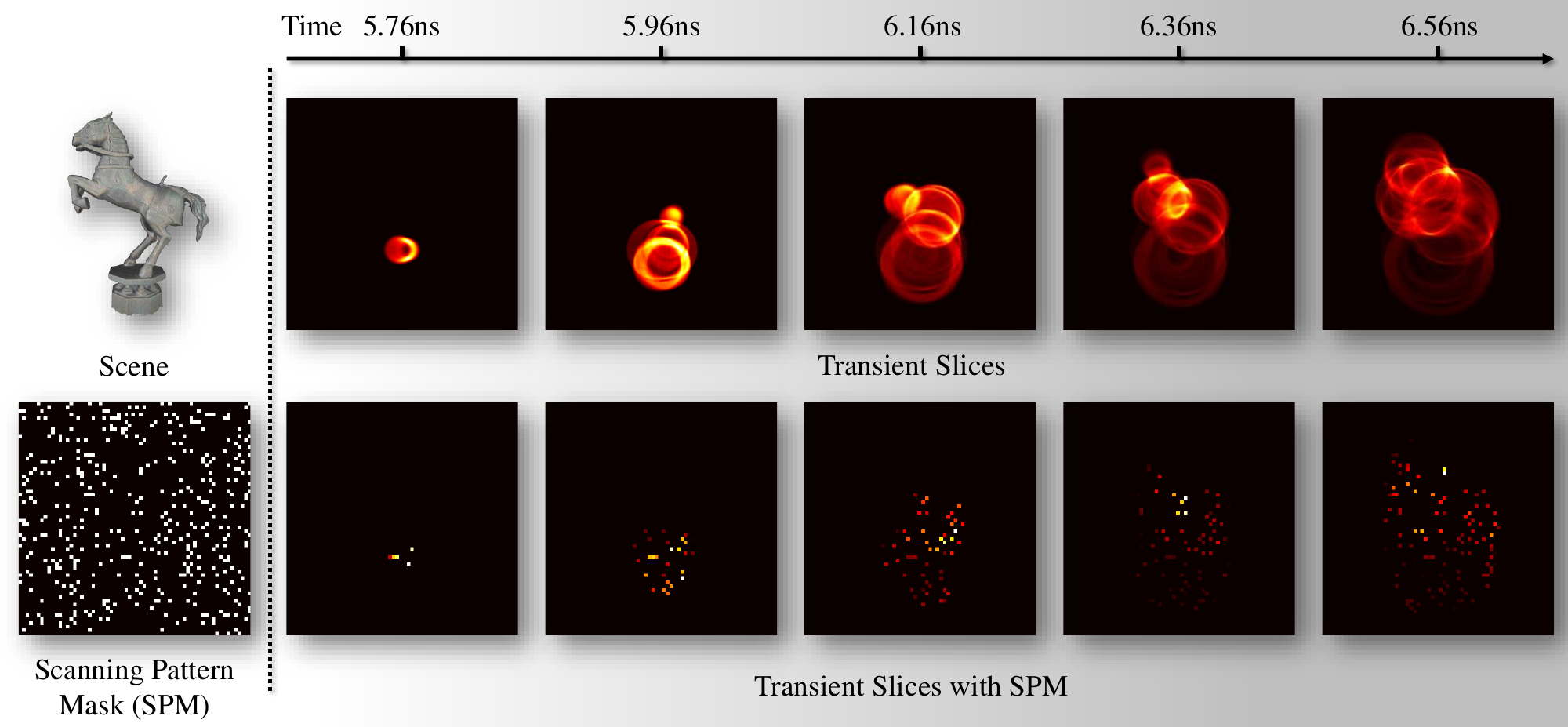}
\end{center}
  \caption{\textbf{Visualization of NLOS transients with and without SPM.} The NLOS transients of a hidden horse are sliced at different arrival times. Using a scanning pattern mask (SPM), the corresponding transients remain as shown in the second row. }
\label{fig:transients}
\end{figure*}


Pretrained models have demonstrated formidable capabilities across multiple domains. Vision encoders ~\cite{simonyan2014vgg, he2016resnet} pretrained on extensive image datasets, like ImageNet-1K~\cite{nips2012imagenet}, have demonstrated the capability to learn latent representations beneficial for various vision tasks, including image classification, object detection, and segmentation. Vision Transformer~\cite{dosovitskiy2021an} employs attention mechanism to process images as sequences of patches. 
Masked modeling is a widely used technique, e.g., BERT~\cite{devlin2019Bert}, BEiT~\cite{bao2022Beit}, BEVT~\cite{wang2022bevt} and GPT~\cite{ChatGPT} in natural language processing. In computer vision, Masked Autoencoder~\cite{he2022MAE} serves as a self-supervised technique by masking a large portion of input data and training the model to recover the original input. This process enhances the model's ability to learn comprehensive latent representations. For video data, MAE has been adapted to include temporal dimensions through cube masking, optimizing the exploitation of temporal information ~\cite{Tong2022videomae, wang2023videomae2, he2022MAEst}. 
We adopt the masked autoencoder as the core of MARMOT for two main reasons. First, NLOS transients are 3D signals with high computational demands; masking reduces training cost by leveraging redundancy. Second, the masking strategy enables flexible simulation of diverse NLOS configurations, allowing MARMOT to generalize across varying input formats and scanning patterns. 

\section{Pretrain MARMOT}
\label{sec:nlosmae}

\subsection{MARMOT Architecture}
\begin{figure*}[t]
\begin{center}
\includegraphics[width=1.0\linewidth]{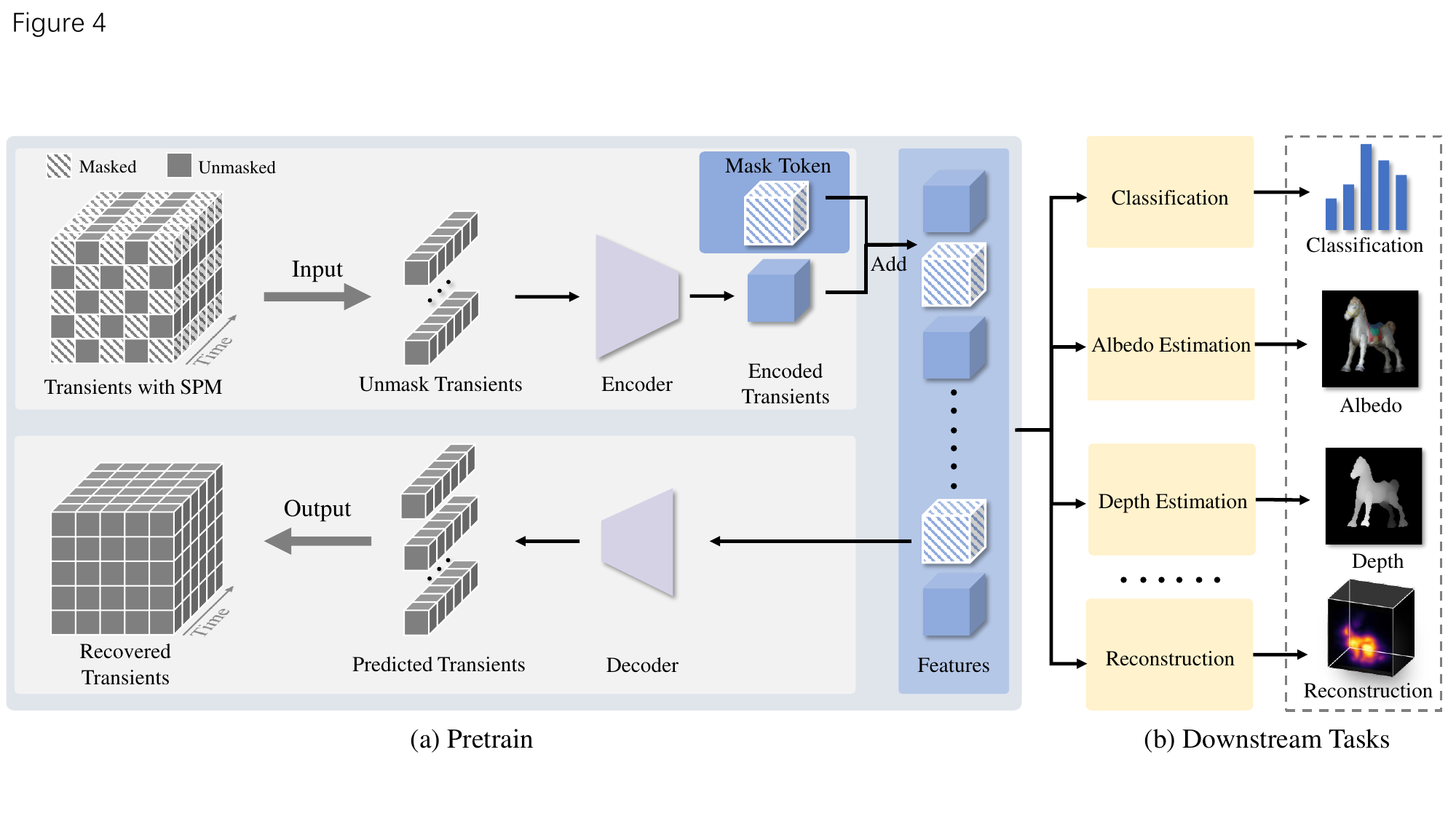}
\end{center}
  \caption{\textbf{Pipeline of MARMOT.} 
(a) MARMOT consists of a pair of Transformer-based encoder and decoder. 
On original transients, we mask random transients and exploit the unmasked subset as input. The encoder estimates latent features, from which the decoder recovers missing transients and combines the unmasked transients to obtain overall transients. (b) MARMOT can facilitate the features from the encoder and the recovered transients to a variety of downstream NLOS tasks.
}
\label{fig:pipeline}
\end{figure*}

MARMOT is a pretrained model designed to handle regularly or irregularly scanned  
and sparse NLOS transients. We show the architecture of MARMOT in Fig.~\ref{fig:pipeline}. Key components of MARMOT include transient masking, encoder, and decoder.

\paragraph{Transient Masking} In a confocal setup, NLOS transients form 3D spatial-temporal signals captured by scanning a relay surface. We introduce a scanning pattern mask (SPM) to mask random transients, mimicking arbitrary scanning patterns, either regular or irregular. Fig.~\ref{fig:transients} visualizes the transients at different arrival times. 
The appropriate masking ratio varies across different data modalities, due to distinct information densities. For instance, languages are well-organized, information-dense signals. Similar to masked autoencoders in vision~\cite{he2022MAE, Tong2022videomae}, which leverage spatial redundancy in images and videos, we observe strong redundancy in transients. Thus, In our experiments, we randomly mask $95\%$ of the origin transients. 
Random masking and high masking ratios of SPM allow MARMOT to flexibly augment data in various NLOS scenarios. More importantly, high masking ratios enable model to process only a small portion of transients, reducing overall training time and resources.   

\paragraph{Encoder} MARMOT employs a transformer-based (ViT~\cite{dosovitskiy2021vit}) encoder to transform the set of unmasked transients into latent features, similar to MAE~\cite{he2022MAE}. Through a series of Transformer blocks with added positional embeddings, MARMOT captures complex relationships between transients. 
A significant challenge in training is the computation and memory cost for training, which substantially increase with the number of measurement points. MARMOT eases this issue by only operating the unmasked subset of the transients (e.g., $5\%$), rather than the full set. 

\paragraph{Decoder} MARMOT exploits a small transformer-based decoder to recover the full set of transients from the encoded tokens (or features) from the encoder and the mask token. Mask token is a learned vector. We combine the recovered and unmasked transients to obtain the full set of complete transients. Recently, several works attempt to recover dense transients from a small number of measurements~\cite{liu2023NC, wang2023non, li2023deep}. 
They only support sparse scanning under uniform distributions.
From our decoder, the recovered transients can be applied to a wide range of NLOS tasks (Sec.~\ref{sec:decoderadapt}).


\subsection{TransVerse dataset}
We generate a large-scale synthetic transient dataset, TransVerse, for NLOS imaging. Fig.~\ref{fig:datasets} shows a confocal NLOS setting (a), scanning points and transients (b), and partial examples of objects in the dataset (c). 
We collect a diverse set of 3D models from Objaverse~\cite{deitke2023objaverse}, ranging from real-world to cartoon-styled objects with simple or complex geometry.  
We simulate confocal NLOS measurements using a GL-based transient renderer~\cite{chen2020learned}. Each object is randomly placed, scaled, and rotated. Transients are rendered at $256 \times 256$ scanning points across a 2m$\times$2m relay surface, with a temporal resolution of 32 picoseconds. 
Fig.~\ref{fig:transients} illustrates examples of simulated transients.
To model realistic noise, we simulate SPAD measurements following~\cite{pan2022onsite}. The final transients $\tau^{\text{SPAD}}$ is:
 \begin{equation}
    \tau^{\text{SPAD}} = \text{Poisson}((\tau*j)+b)
    \label{eq:spad_model}
\end{equation}
\noindent where $\tau$ and $\tau^{\text{SPAD}}$ represent the synthesized transients and the transients with SPAD-based noise. $j$ is the temporal jitter of the whole system. The bias $b$ is introduced due to ambient lights and dark counts of the SPAD and is considered to be independent of time at scanning points. 




\subsection{Pretrain MARMOT with TransVerse}

With TransVerse, we pretrain MARMOT in a self-supervised manner, as illustrated in Fig.~\ref{fig:pipeline}. We mask random the complete transients and input the unmasked subset to the encoder. Masked token is appended to the output latent features, and the decoder recovers the missing transients. We optimize MARMOT by minimizing a mean squared error loss:

\begin{equation}
    \mathcal{L} = \frac{1}{N} \sum_{i=1}^{N} (\tau_i - \hat{\tau}_i)^2
    \label{eq:loss}
\end{equation}

\noindent where $N$ is the number of transients. $\tau_i$ denotes the transients at point $i$ and $\hat{\tau}_i$ represents the predicted transients. Following MAE~\cite{he2022MAE}, we only exploit the subset of masked transients, rather than the full set, to compute the loss. The overall self-supervised learning strategy enables MARMOT to leverage intrinsic features convolved in the NLOS transients. 
Detailed architecture and training setups are provided in Supplementary Material Sec.~\ref{sec:training}.

\begin{figure}
\centering
\includegraphics[width=1.0\linewidth]{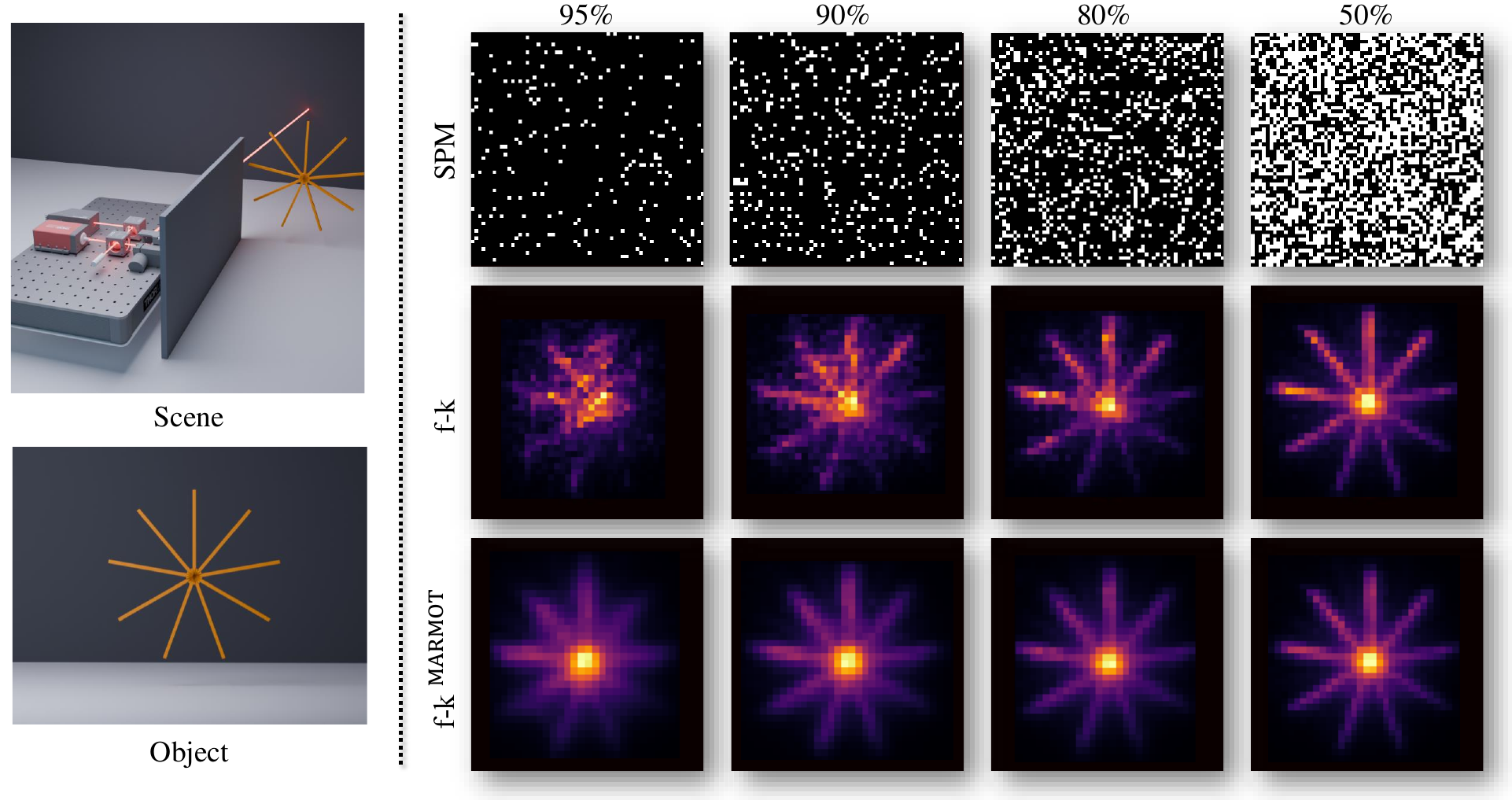}
  \caption{\textbf{Reconstruction results on different masking ratio.} We randomly masked $50\%, 80\%, 90\%,$ and $95\%$ of the complete transients using corresponding SPMs. We display the results of f-k applied to the masked transients and the transients predicted by MARMOT.}
\label{fig:ratio}
\end{figure}

\section{NLOS reconstruction} 
\label{sec:decoderadapt}
MARMOT recovers missing transients from the latent features estimated in the encoder. This process allows MARMOT to tackle challenging NLOS tasks, such as transient completion and shape recovery.


\subsection{Transient Completion}
MARMOT's decoder predicts missing transients from output features of the encoder. 
We provide the visualization of recovered transients in Supplementary Material Fig.~\ref{fig:transients2}.

We first demonstrate that, although MARMOT is trained with a masking ratio of $95\%$, it is robust to different masking ratios. Fig.~\ref{fig:ratio} shows the reconstruction results of MARMOT under different masking ratios using f-k~\cite{lindell2019fk}. 
Since f-k requires full resolution input, we first apply interpolation to the masked measurements.
As the masking ratio increases, the results of f-k gradually become fragmented until the shape of the object is indistinguishable. In contrast, MARMOT produces complete shape of the object with clear edges even at $95\%$ masking ratio.

\begin{figure}[t]
\begin{center}
\includegraphics[width=0.83\linewidth]{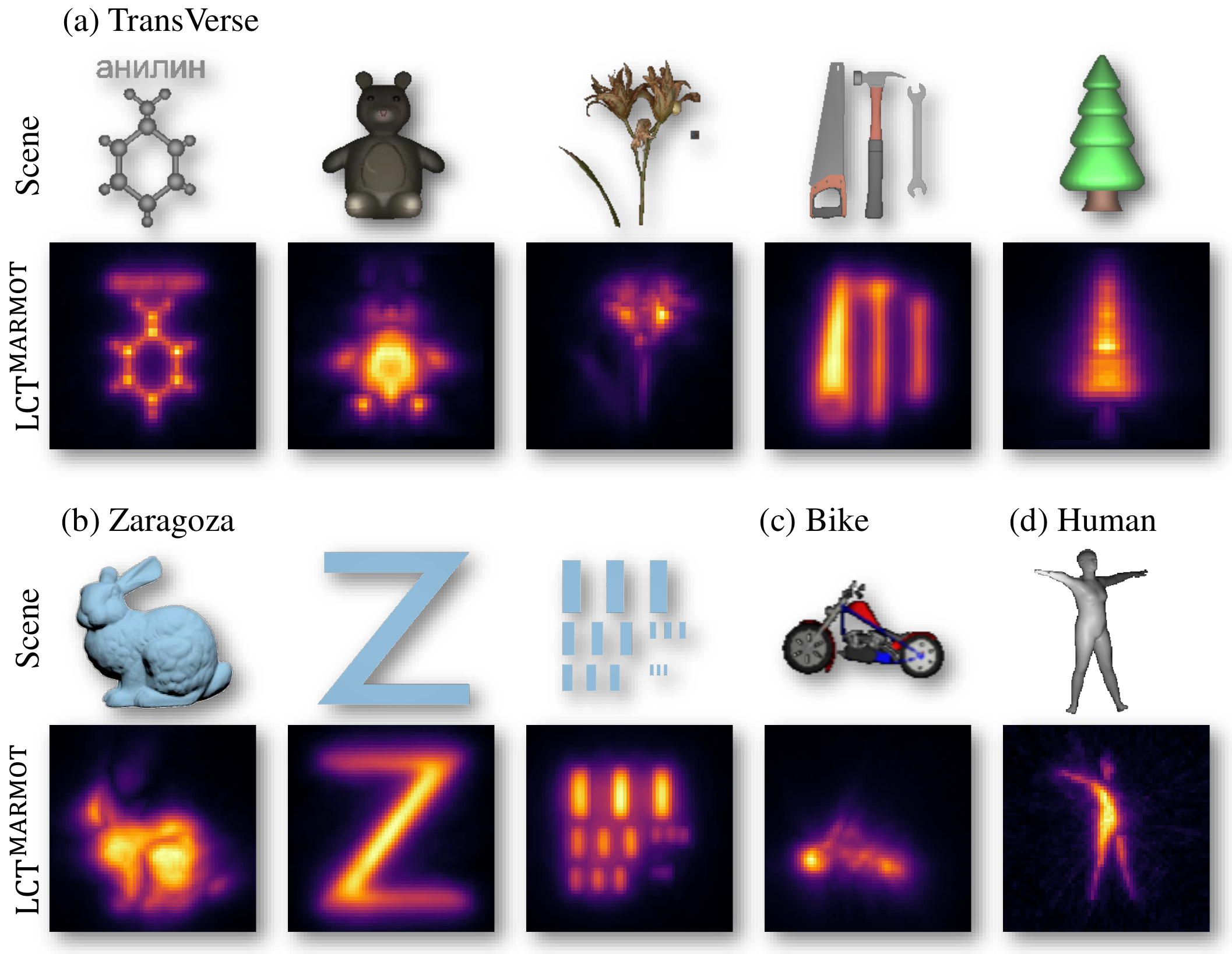}
\end{center}
\caption{\textbf{Reconstruction results on different datasets.} 
We select various hidden objects from different datasets, including TransVerse, Zaragoza~\cite{galindo19-NLOSDataset-zaragoza}, Bike~\cite{chen2020learned}, and Human~\cite{yu2023enhancing}. We then mask $95\%$ of the transients for each object under random SPMs and use MARMOT to predict the full transients from the masked input. Finally, we reconstruct the objects using LCT~\cite{otoole2018LCT}.}

\label{fig:recon}
\end{figure}

MARMOT also demonstrates robustness across different datasets. We select synthetic transients from four datasets: TransVerse, Zaragoza~\cite{galindo19-NLOSDataset-zaragoza}, Bike~\cite{chen2020learned}, and Human~\cite{yu2023enhancing}. The objects have varieties of geometry and appearance. During pretraining, we use transients from TransVerse only, dowsampling to $64\times 64$. We mask random transients with $95\%$ of the full set and recover the overall transients. 
Using the recovered transients, we reconstruct the hidden objects using the physics-based LCT~\cite{otoole2018LCT}.
In addition, we adapt MARMOT to other three datasets (Zaragoza~\cite{galindo19-NLOSDataset-zaragoza}, Bike~\cite{chen2020learned}, and Human~\cite{yu2023enhancing}), and evaluate the generalization capacity of MARMOT. 
The results shown in Fig.~\ref{fig:recon} and Supplementary Material Sec.~\ref{sec:morenlos} demonstrate that MARMOT is generalizable to different datasets.

\begin{figure}[t]
\begin{center}
\includegraphics[width=0.99\linewidth]{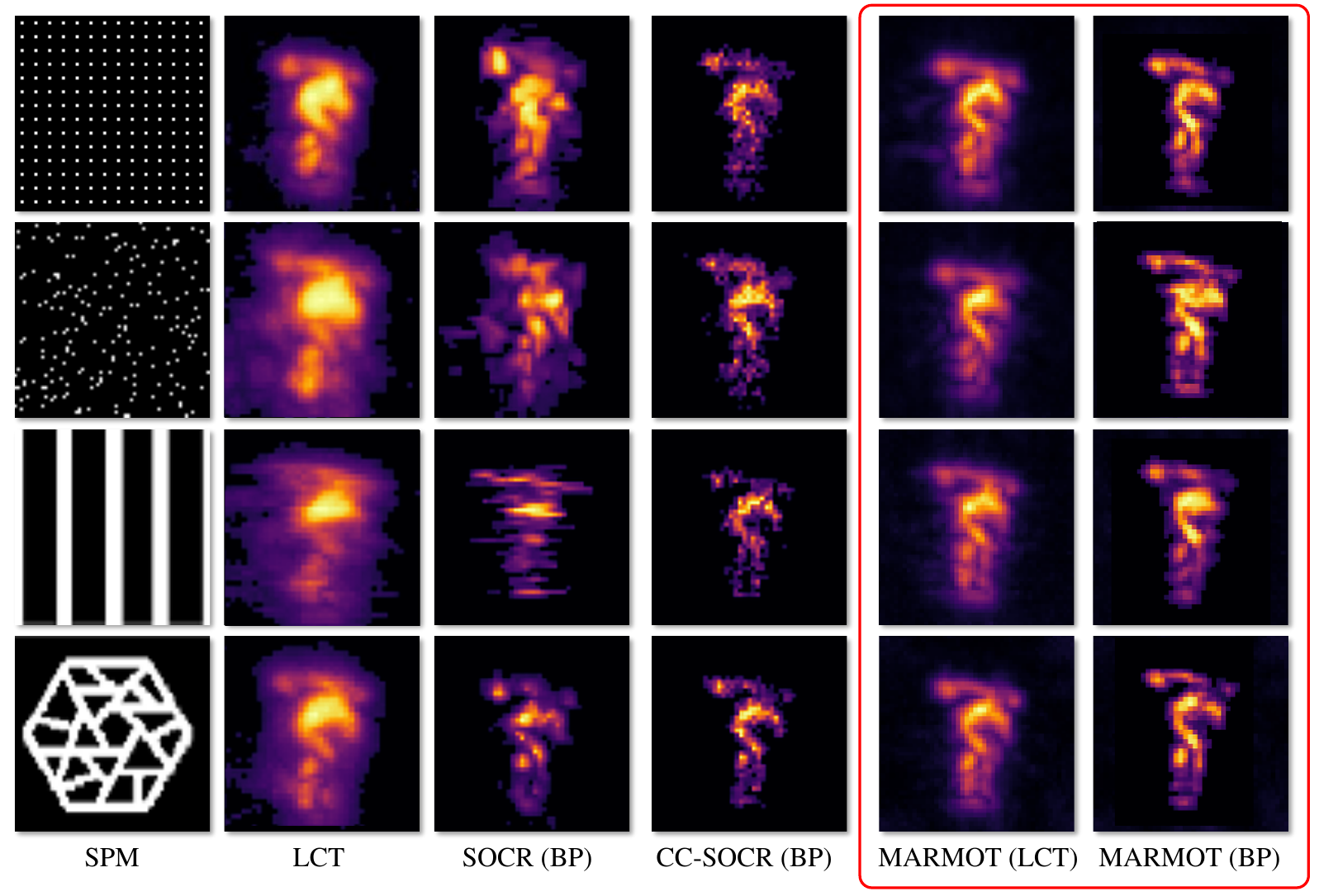}
\end{center}
  \caption{\textbf{Comparison between MARMOT and baseline methods on real-measured dataset.} We exploit the transients of a hidden statue from the Stanford f-k dataset~\cite{lindell2019fk}. We mask the complete transients using four distinct SPMs and recover via our MARMOT and baseline methods: SOCR~\cite{liu2021non}, CC-SOCR~\cite{liu2023NC}. The second column shows the results using LCT from masked transients. }
\label{fig:comparereal}
\end{figure}

We also compare MARMOT with state-of-the-art methods designed for sparse and irregular transients, including SOCR~\cite{liu2021non} and CC-SOCR~\cite{liu2023NC}. Unlike these approaches, which directly complete transients via interpolation and optimization, MARMOT reconstructs transients in latent space instead. We conduct experiments on real data from the f-k dataset~\cite{lindell2019fk} under four distinct scanning pattern masks (SPMs), representing dense/sparse and regular/irregular sampling.
Fig.~\ref{fig:comparereal} shows the qualitative results.
While LCT fails to reconstruct recognizable geometry, and SOCR/CC-SOCR recover blurred shapes via BP~\cite{velten2012NC}, our method enables sharper reconstruction using both LCT~\cite{otoole2018LCT} and BP. We also provide more qualitative and quantitative results in Supplementary Material Sec.~\ref{sec:morenlos}.
MARMOT consistently outperforms baselines. Its robustness highlights its potential in diverse real-world NLOS applications. 

\subsection{Shape Recovery}
\begin{wrapfigure}{r}{0.40\textwidth}
  \centering
  \vspace{-2em}
  \includegraphics[width=0.38\textwidth]{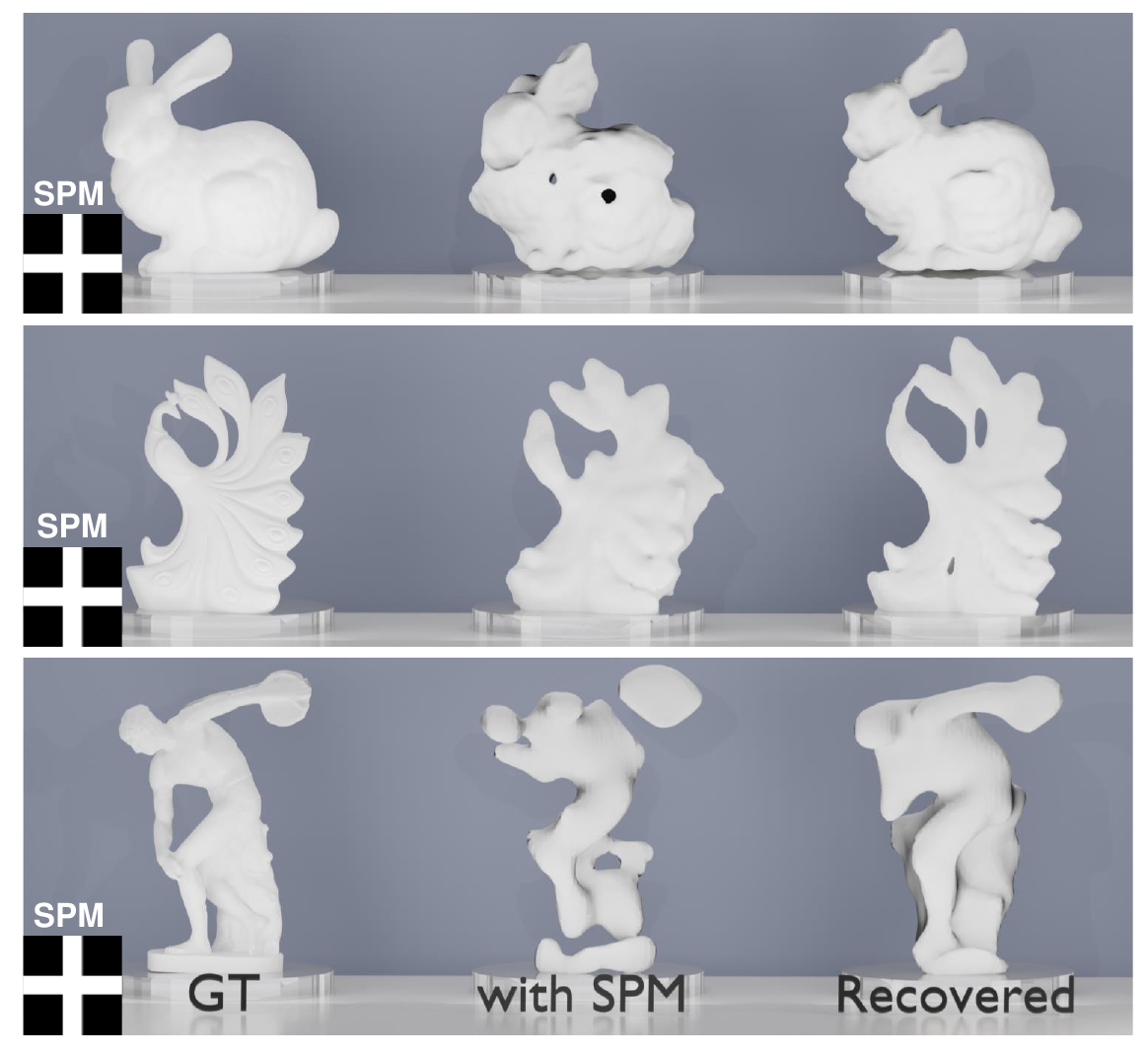}
  \vspace{-10pt}
      \caption{\textbf{3D reconstruction results.} We mask transients with SPM and recover them via MARMOT, then reconstruct 3D models using a neural-field method~\cite{huang2023omni}.}
    \label{fig:3drecon}
    \vspace{-6em}
\end{wrapfigure}

3D NLOS reconstruction is challenging because we need to resolve the inverse problem from complexly integrated signals. Early NLOS imaging methods use volumetric representation and need extra shape extraction algorithms~\cite{velten2012NC,otoole2018LCT,lindell2019fk,liu2019PF}. Several approaches employ geometric representation for direct shape optimization~\cite{xin2019theory,tsai2019beyond}. Recent neural fields utilize neural networks as implicit representations to extract geometric information~\cite{shen2021NeTF,fujimura2023nlosneus}. 
MARMOT offers a useful alternative to predict transients from sparse sampling and expend the potentials for 3D NLOS reconstruction. 
Fig.~\ref{fig:3drecon} showcases 3D reconstruction of three hidden objects. Using a SPM, we mask $95\%$ of the complete transients at spatial resolution of $64\times64$. We exploit the neural-field method~\cite{huang2023omni}, which supports both dense and sparse transients as input, to reconstruct 3D models of these objects using unmasked (i.e., sparse) and recovered transients. We compare the results to ground truths (GT), which are rendered using original meshes of the objects. 

In summary, extensive experimental results for NLOS reconstruction validate that MARMOT can accurately predict the complete set of transients with arbitrary masking patterns and high masking ratios such as $95\%$. This enables us to scale up MARMOT to a pretrained model for transient imaging. 

\section{Reconstruction-free Visual Inference}
\label{sec:finetune}

MARMOT can not only be applied in NLOS reconstruction, but also to different visual tasks in a different paradigm. Using the pretrained encoder, we transfer the learned features to three downstream tasks, including albedo estimation, depth estimation, and classification.

\paragraph{Baseline Methods} We compare our results with two SOTA learning-based methods: NLOST~\cite{li2023nlost}, AGK~\cite{yu2023enhancing} and three physics-based methods: LCT~\cite{otoole2018LCT}, f-k~\cite{lindell2019fk} and PF~\cite{liu2019PF}. NLOST employs a Transformer-based architecture, while AGK utilizes an attention-guided end-to-end neural network; both methods are designed to process uniformly sampled and densely measured transients. Although recent learning-based approaches~\cite{li2023deep, wang2023non} have shown progress in handling sparse transients, they still rely on inputs collected over regular grids.

\subsection{Classification}

\begin{wraptable}{r}{0.4\textwidth}
\label{tab:classification}
  \vspace{-2em}
\centering
\caption{\textbf{Classification comparison.} Precision / Recall / Accuracy metrics.}
\small
\begin{tabular}{cccc}
\toprule
\textbf{Methods} & \textbf{Prec}$\uparrow$ & \textbf{Rec}$\uparrow$ & \textbf{Acc}$\uparrow$ \\
\midrule
NLOST  & 0.93 & 0.93 & 0.91 \\ 
AGK    & 0.95 & \textbf{0.94} & 0.94 \\ 
MARMOT & \textbf{0.96} & \textbf{0.94} & \textbf{0.95} \\ 
\bottomrule
\end{tabular}
\end{wraptable}

We evaluate the classification performance of MARMOT on a publicly available dataset~\cite{yu2023enhancing, mu2022NLOS3d}, which contains transient measurements of 2,642 objects, including Arabic digits, as well as English and Roman letters. To ensure fair comparison, we use transients at a spatial resolution of $64 \times 64$ for all methods. For MARMOT, we append pooling and linear layers to the pretrained encoder and finetune only the decoder for 10 epochs. Classification performance is evaluated using precision, recall, and accuracy metrics, as summarized in Table~\ref{tab:classification}.


\subsection{Albedo Estimation}

\begin{figure}[t]
\begin{center}
\includegraphics[width=\linewidth]{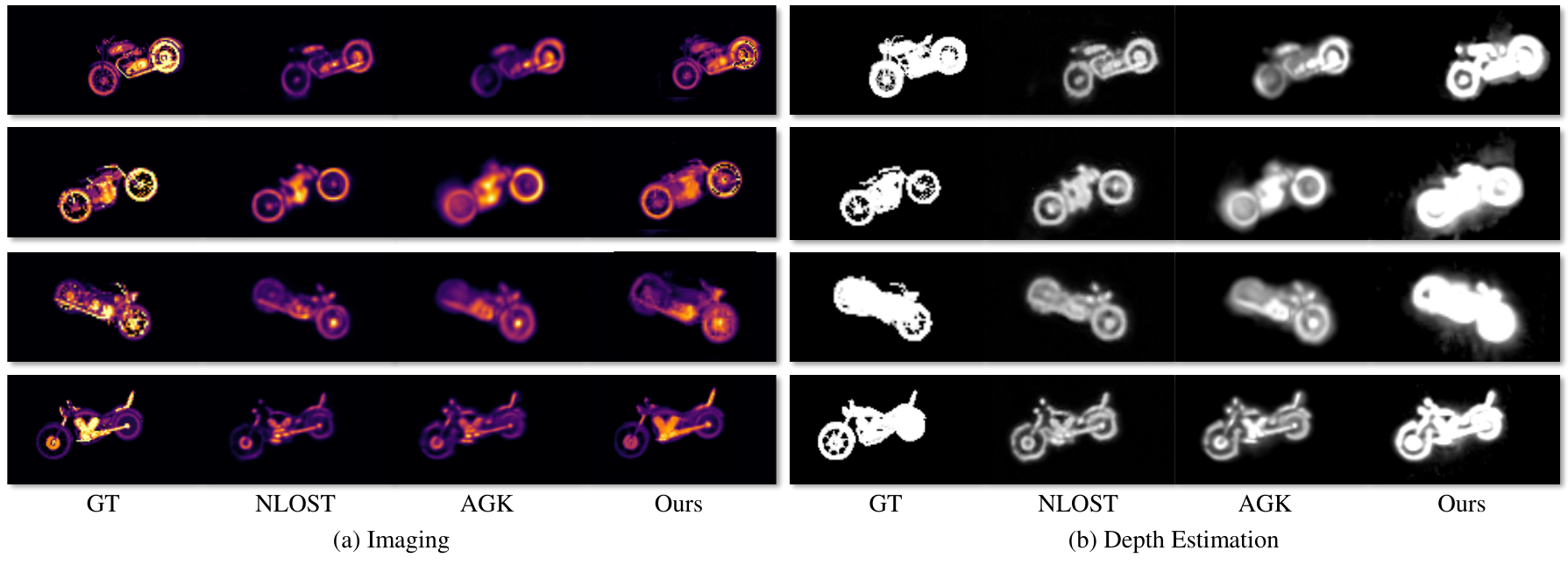}
\end{center}
  \caption{\textbf{Qualitative comparison for albedo / depth estimation task.} We compare MARMOT with NLOST~\cite{li2023nlost} and AGK~\cite{yu2023enhancing}. GTs are rendered using original meshes of motorbikes. }
\label{fig:imaging}
\end{figure}


Estimating albedo of hidden objects from transients is challenging since the information is heavily convolved with multiple light paths. One needs to solve the inverse imaging problem and reconstruct the hidden objects from complex integral signals. We adapt our pretrained encoder to albedo estimation on the motorbike datasets~\cite{chen2020learned} and use spatial resolution of $128 \times 128$ as input. 
Previous methods directly employ CNN-based decoders to reconstruct images from features. In contrast, we adopt an advanced diffusion model approach~\cite{su2024model}. We use the features extracted by MARMOT as conditional inputs to guide the diffusion process. The detailed implementation is provided in Supplementary Materials. We train the diffusion pipeline to estimate albedo for 1000 epochs.

Fig.~\ref{fig:imaging}(a) showcases estimated albedo images of hidden motorbikes using the full set of transients for three methods. The images from NLOST and AGK are blurry whereas our images from MARMOT show clear and complex structures of the motorbikes, e.g., the wheels. 
Table~\ref{tab:imaging} shows the quantitative results. We provide more comparison in Supplementary Materiel Sec.~\ref{sec:morenlos}.

\begin{table}[t]
\centering
\begin{minipage}[ht]{0.5\textwidth}
\centering
\small
\setlength{\tabcolsep}{2pt}
\label{tab:imaging}
\caption{Quantitative comparison for Fig.~\ref{fig:imaging}(a)}
\resizebox{0.8\linewidth}{!}{
\begin{tabular}{cccccc}
\toprule
\textbf{Task}   & \textbf{Methods}  & \textbf{ED$\downarrow$} & \textbf{CS$\uparrow$} & \textbf{SSIM$\uparrow$} & \textbf{PSNR$\uparrow$} \\ \midrule
\multirow{3}{*}{Albedo} & NLOST  & 0.0614 & 0.9293 & 0.9499 & 24.56\\
                        & AGK    & 0.0642 & 0.9144 & 0.9436 & 24.73\\
                        & MARMOT & \textbf{0.0578} & \textbf{0.9405} & \textbf{0.9573} & \textbf{25.04}\\ 
                        \bottomrule
\end{tabular}}
\end{minipage}%
\hfill
\begin{minipage}[ht]{0.5\textwidth}
\centering
\small
\setlength{\tabcolsep}{2pt}
\label{tab:depth}
\caption{Quantitative comparison for Fig.~\ref{fig:imaging}(b)}
\resizebox{0.8\linewidth}{!}{
\begin{tabular}{cccccc}
\toprule
\textbf{Task}   & \textbf{Methods}  & \textbf{ED$\downarrow$} & \textbf{CS$\uparrow$} & \textbf{SSIM$\uparrow$} & \textbf{PSNR$\uparrow$} \\ \midrule
\multirow{3}{*}{Depth} & NLOST & 0.1097 & \textbf{0.9277} & 0.7631 & 19.52\\
                       & AGK   & 0.0956 & 0.9141 & 0.9094 & 20.02\\
                       & MARMOT & \textbf{0.0899} & 0.8982 & \textbf{0.9117} & \textbf{20.29}\\ 
                       \bottomrule
\end{tabular}}
\end{minipage}
\vspace{-10pt}
\end{table}

\subsection{Depth Estimation}

On the same motorbike dataset, we facilitate MARMOT to depth estimation. This task aims to recover hidden objects with detailed geometries to distinguish precise location and structure. Our pipeline for depth estimation is same as albedo estimation because both tasks reconstruct a 2D map of the hidden object. We finetune the depth estimation decoder for 1000 epochs. Fig.~\ref{fig:imaging}(b) and Table~\ref{tab:depth} show the depth maps of hidden motorbikes, MARMOT outperforms the baseline methods as well.

In summary, the above experimental results demonstrate that our pretrained encoder extracts coarse and fine features from sparse transients, enabling detailed hidden-object geometry recovery and generalizable NLOS imaging for diverse applications through finetuning.

\section{Discussion and Conclusion}
\label{sec:conclusion}


In this work, we propose MARMOT, a masked autoencoder for transient imaging particularly in non-line-of-sight (NLOS) scenarios. Our self-supervised model demonstrates strong performance in transients completion and 3D reconstruction, enabling physics/learning-based methods to handle sparse, irregular transients. We have carried out comprehensive experiments and verified the robustness and capibility of MARMOT. Furthermore, our large-scale dataset, TransVerse, contains 500K objects and will benefit the NLOS research community.

While MARMOT offers an alternative solution to the inverse NLOS imaging problem, there remain following challenges. First, NLOS transients are supposed to be captured on a planar surface; otherwise, we need to pre-calibrate the scanning surface and correct the transients as the prior work~\cite{lindell2019fk}. Second, we consider only the confocal NLOS configuration, though we can process transients under non-confocal settings via a transforming algorithm~\cite{lindell2019fk,liu2021non}, this may introduce additional error. Finally, our TransVerse dataset has not yet contained objects with complex materials such as specular. We consider these points as interesting direction of future work. MARMOT offers a meaningful alternative to model transient imaging, specifically NLOS imaging, and will potentially motivate solutions for other challenging inverse problems. 

\bibliographystyle{IEEEtran}

\bibliography{neurips_2025}

\newpage
\appendix

\section*{Supplementary Material for MARMOT: Masked Autoencoder for Modeling Transient Imaging}

\section{Transients Completion}
\begin{figure*}[t]
\begin{center}
\includegraphics[width=0.99\linewidth]{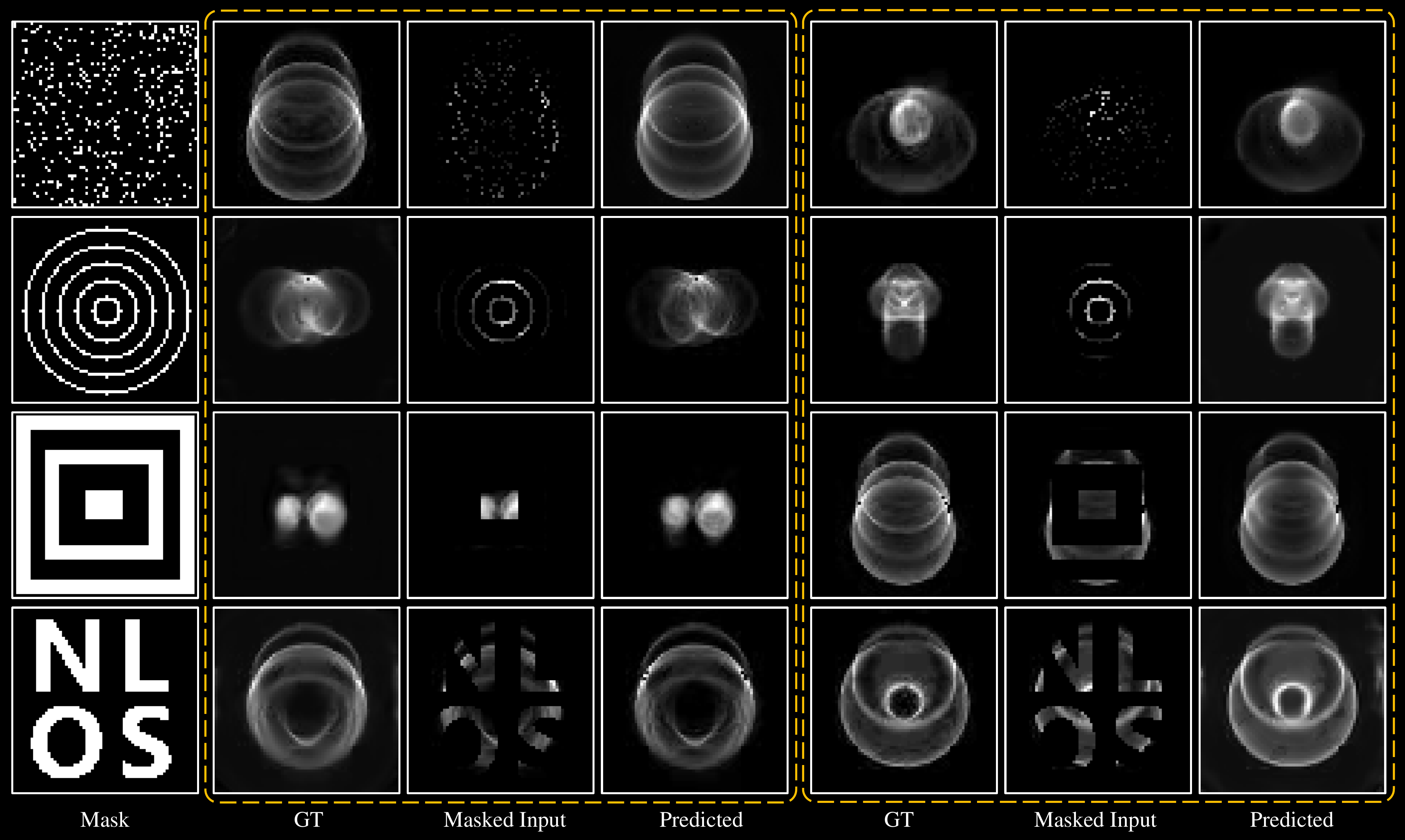}
\end{center}
  \caption{\textbf{Visualization of origin and recovered transients.} We display the original fully sampled transients slices alongside the transients slices recovered using our method. Each group of slices was masked according to the pattern shown in the first column and then reconstructed. 
}
\label{fig:transients2}
\end{figure*}

We visualize the recovered and origin transeints in Fig.~\ref{fig:transients2}. All the transients as input has resolution of $64 \times 64$ and masked using the pattern in the first column.
It's notable that the model in trained using random masking of masking ratio $95\%$ and after pretraining it can be used under different scanning patterns.
By visually comparing the original and reconstructed transients, one can appreciate the fidelity with which our method restores missing information, demonstrating the potential of our approach in practical NLOS imaging scenarios where data may often be sparse or corrupted. 

\section{Training details}\label{sec:training}
\textbf{Implementation.} We conduct pretraining on TransVerse dataset for 10 epochs using 64 NVIDIA A800 GPUs. This process takes approximately 48 hours. We experiment MARMOT on transients of two spatial-resolution versions: $128 \times 128$ and $64 \times 64$. All experiments are carried out on the PyTorch platform. Table~\ref{tab:structure} summarizes the detailed parameters of MARMOT architecture. Because the intensity of transients depends on the laser's average power and the exposure time, we normalize each transient in range of 0 and 1. 

\textbf{Ablation.} We carry out ablation studies to evaluate the impact of MARMOT parameters. Table~\ref{tab:ablations} shows quantitative results by varying depth and width of the encoder and the decoder. 
The depth and the width are defined as the numbers of Transformer blocks and of channels, respectively.
If not specified, the default is: the encoder has depth 24 and width 1024; the decoder has depth 8 and width 512.
Four metrics are exploited to assess the impact, including Euclidean Distance (ED), Cosine Similarity (CS), Structural Similarity Index (SSIM), and Peak Signal-to-Noise Ratio (PSNR). We vary the encoder's depth and width, the scores of four metrics show that 16 or 24 are optimal for the depth, 512 and 1024 for the width, respectively. This implies that the higher-dimensional encoder enables us to scale up the higher capacity of MARMOT. Similarly, optimal decoder's depth is 8, and width is 256 and 512. Our decoder is narrower and shallower than the encoder. This asymmetric design can further accelerate training and produce meaningful features from the network. For experiments, we select the parameters for MARMOT architecture in Table~\ref{tab:structure}. 

\section{More results of NLOS Reconstruction}
\label{sec:morenlos}
\begin{figure*}[t]
\begin{center}
\includegraphics[width=0.99\linewidth]{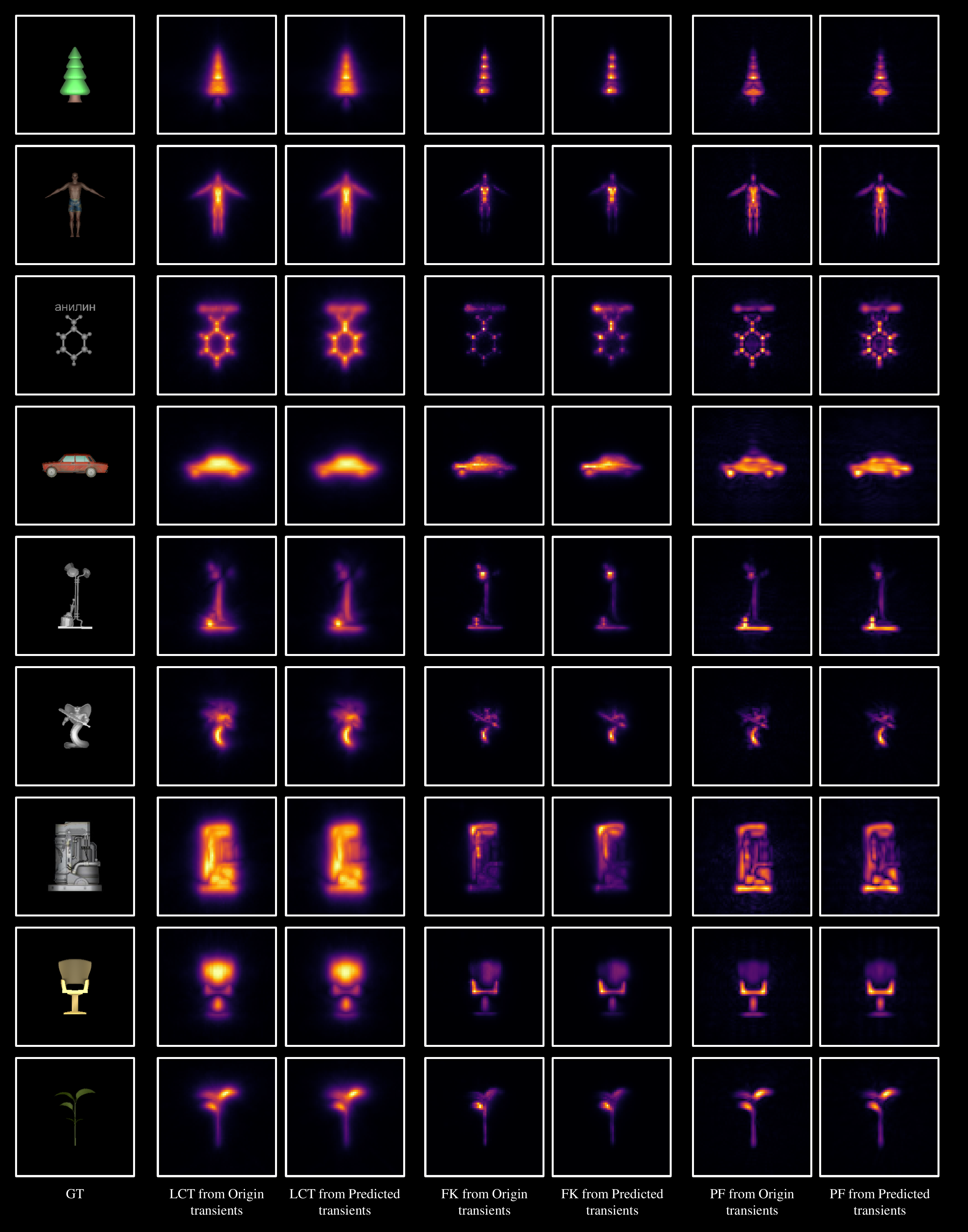}
\end{center}
  \caption{\textbf{Visualization of TransVerse reconstruction results.} We showcase the reconstruction results of transients from the TransVerse dataset. For each set of results, the left side displays reconstructions from the complete original data, while the right side shows reconstructions from transients that were masked by $95\%$ and subsequently restored using our method. For the reconstruction, we employed various methods including LCT, F-K, and PF.
}
\label{fig:nlosverse1}
\end{figure*}

\begin{figure*}[t]
\begin{center}
\includegraphics[width=0.89\linewidth]{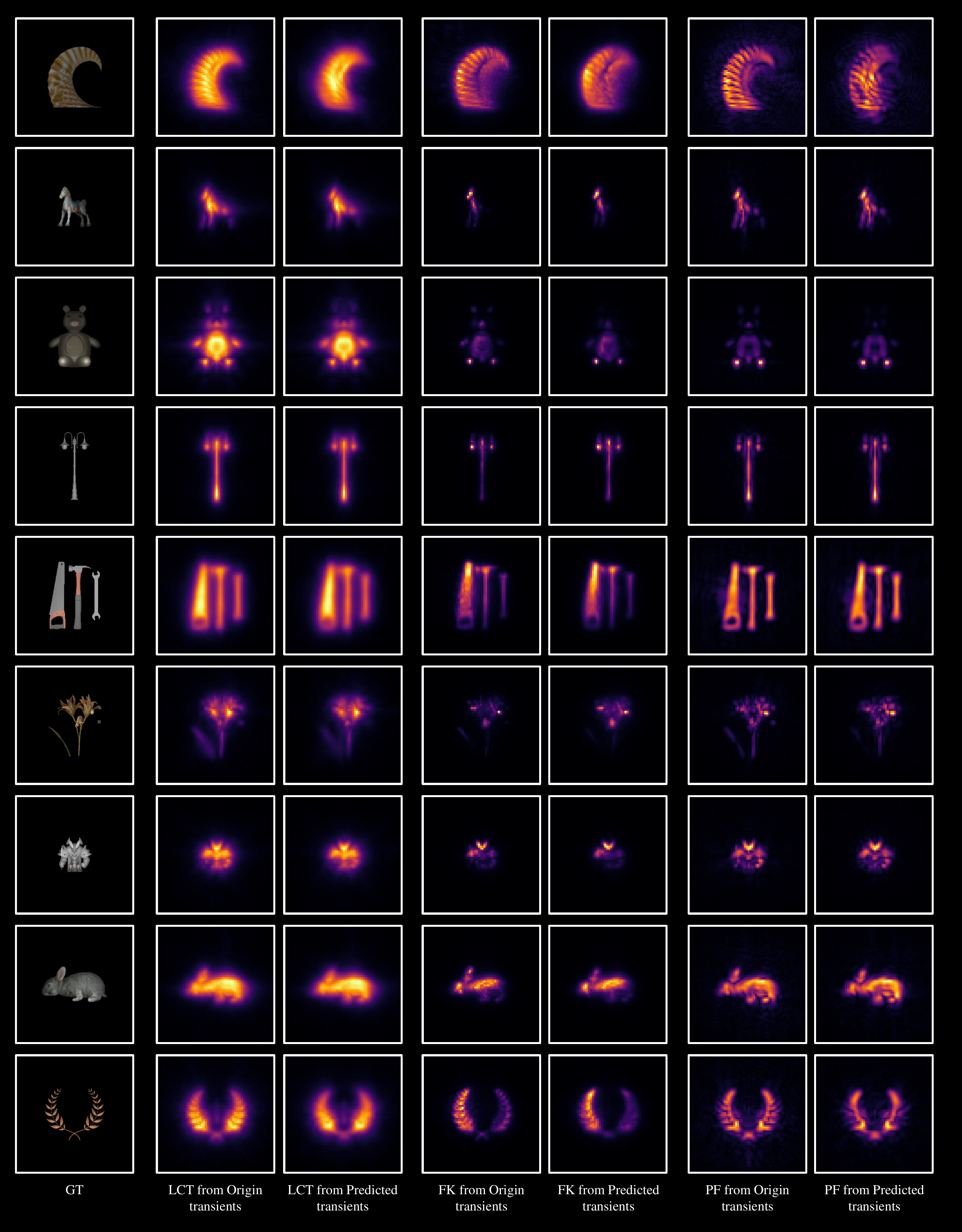}
\end{center}
  \caption{\textbf{Visualization of TransVerse reconstruction results.} We showcase the reconstruction results of transients from the TransVerse dataset. For each set of results, the left side displays reconstructions from the complete original data, while the right side shows reconstructions from transients that were masked by $95\%$ and subsequently restored using our method. For the reconstruction, we employed various methods including LCT, F-K, and PF.
}
\label{fig:nlosverse2}
\end{figure*}

We prvoide more results on TransVerse dataset in Fig.~\ref{fig:nlosverse1} and~\ref{fig:nlosverse2}. We show the reconstruction results from the origin fully sampled and the recovered transients. All the transients as input has resolution of $64 \times 64$ and masked using a random $95\%$ pattern. The reconstrcution methods includes LCT~\cite{otoole2018LCT},f-k~\cite{lindell2019fk} and PF~\cite{liu2019PF}.

\begin{figure*}[t]
\begin{center}
\includegraphics[width=0.8\linewidth]{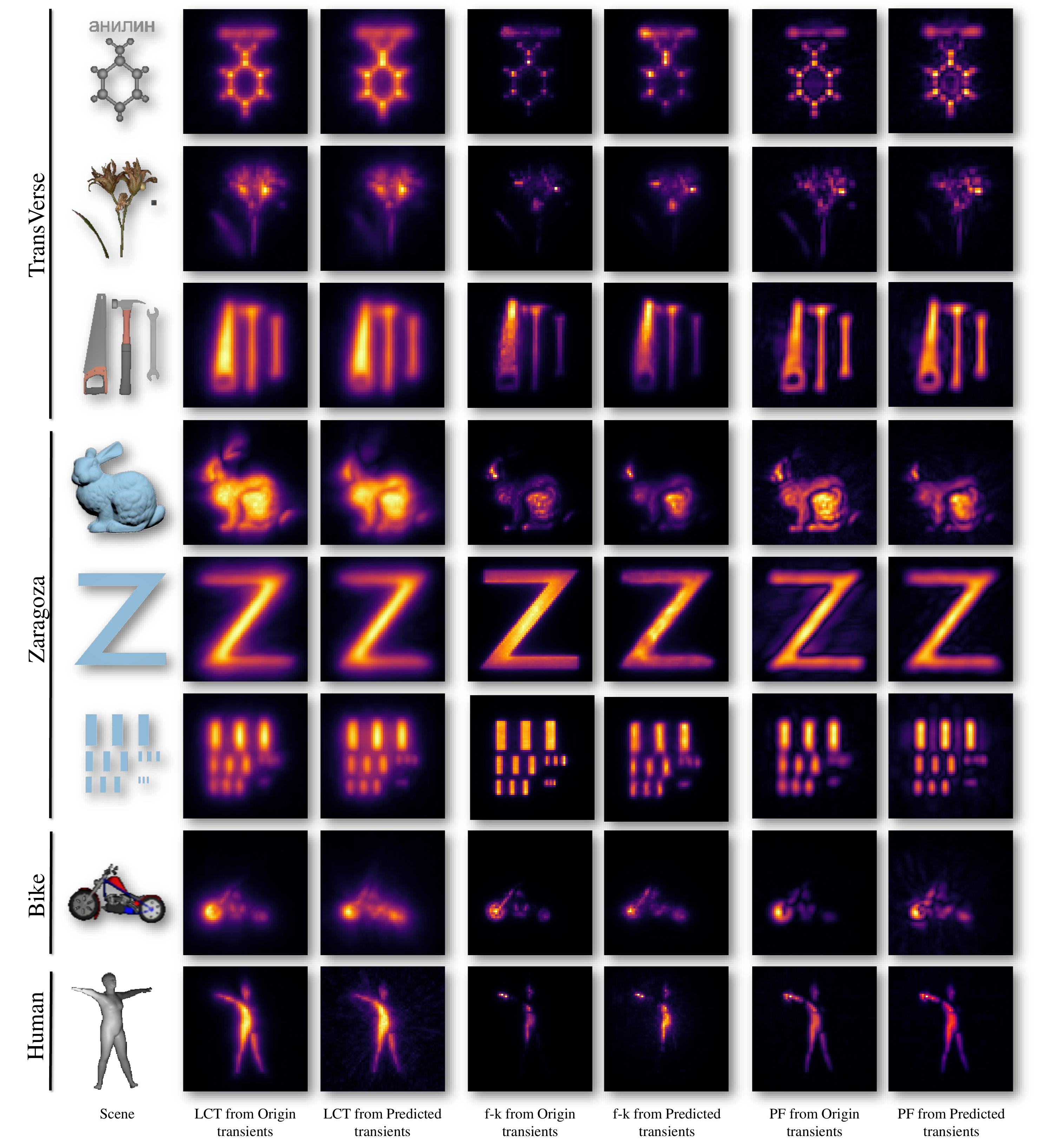}
\end{center}
  \caption{\textbf{Visualization of reconstruction results across different datasets.} We showcase the reconstruction results of transients from the different datasets. For each set of results, the left side displays reconstructions from the complete original data, while the right side shows reconstructions from transients that were masked by $95\%$ and subsequently restored using our method. For the reconstruction, we employed various methods including LCT, F-K, and PF.
}
\label{fig:diffdata}
\end{figure*}

We prvoide more results on different public datasets in Fig.~\ref{fig:diffdata}, including TransVerse, Zaragoza~\cite{galindo19-NLOSDataset-zaragoza}, Bike~\cite{chen2020learned}, and Human~\cite{yu2023enhancing}.. We show the reconstruction results from the origin fully sampled and the recovered transients. All the transients as input has resolution of $64 \times 64$ and masked using a random $95\%$ pattern. The reconstrcution methods includes LCT~\cite{otoole2018LCT},f-k~\cite{lindell2019fk} and PF~\cite{liu2019PF}.

\begin{figure*}[t]
\begin{center}
\includegraphics[width=0.99\linewidth]{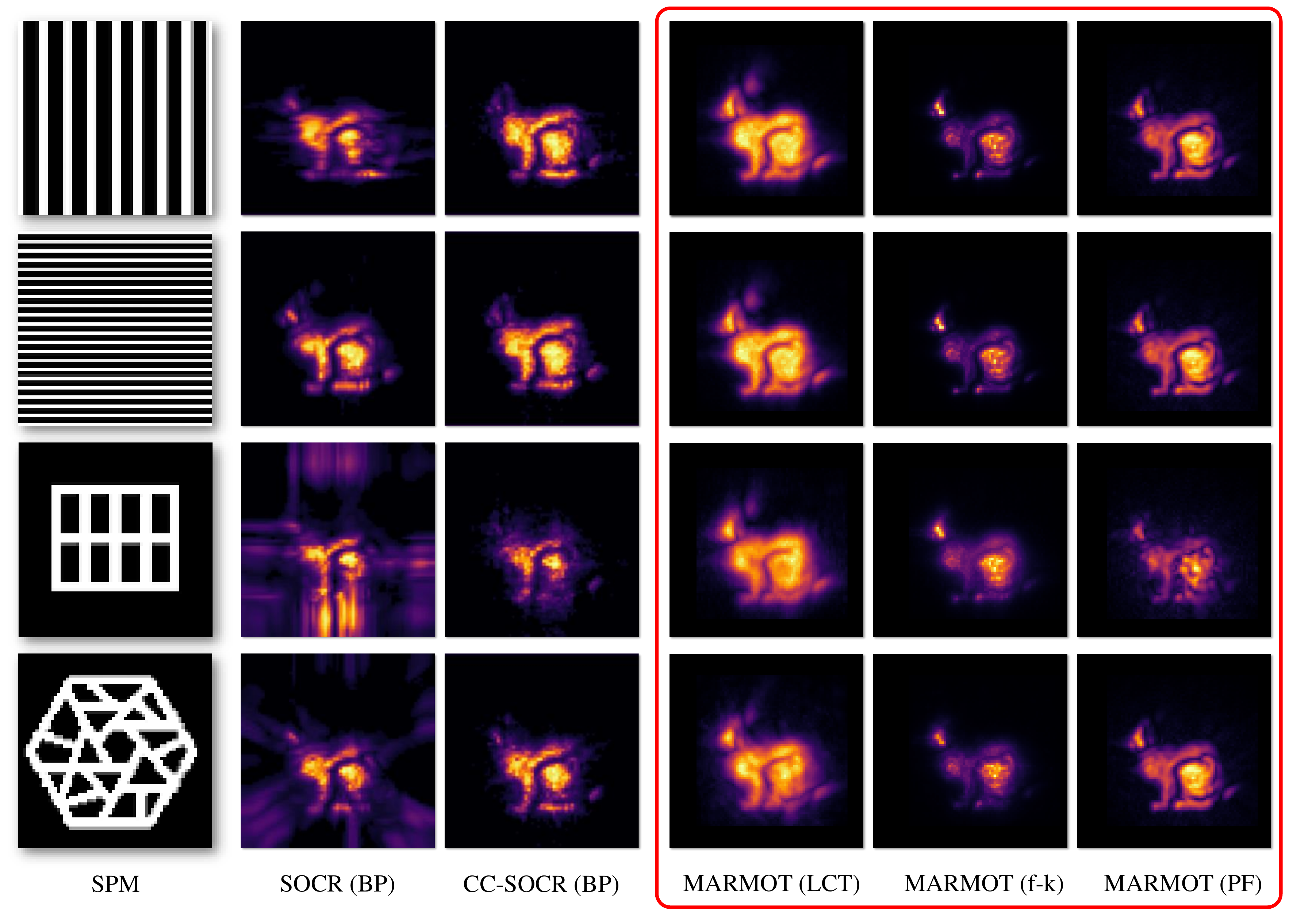}
\end{center}
  \caption{\textbf{Comparison between MARMOT and baseline methods on synthetic dataset.} We exploit the transients of a hidden bunny from the Zaragoza dataset~\cite{galindo19-NLOSDataset-zaragoza}. We mask the original transients using four SPMs and recover the  missing transients using MARMOT and beseline methods: SOCR~\cite{liu2021non} and CC-SOCR~\cite{liu2023NC}. We then exploit overall transients to reconstruct the bunny based on BP~\cite{velten2012NC} for SOCR and CC-SOCR. The results of our MARMOT are based on LCT~\cite{otoole2018LCT}, f-k~\cite{lindell2019fk}, and PF~\cite{liu2019PF}. }
  
\label{fig:comparesyth}
\end{figure*}

We also prvoide more comparison on public dataset in Fig.~\ref{fig:comparesyth}. 
We show the reconstruction results applying LCT on unmasked subset of transients. We also show the results using SOCR~\cite{liu2021non} and CC-SOCR~\cite{liu2023NC} as comparison. Quantitive results are shown in Table.~\ref{tab:comparesyn2}.

\begin{figure}[t]
\vspace{-10pt}
\begin{center}
\includegraphics[width=\linewidth]{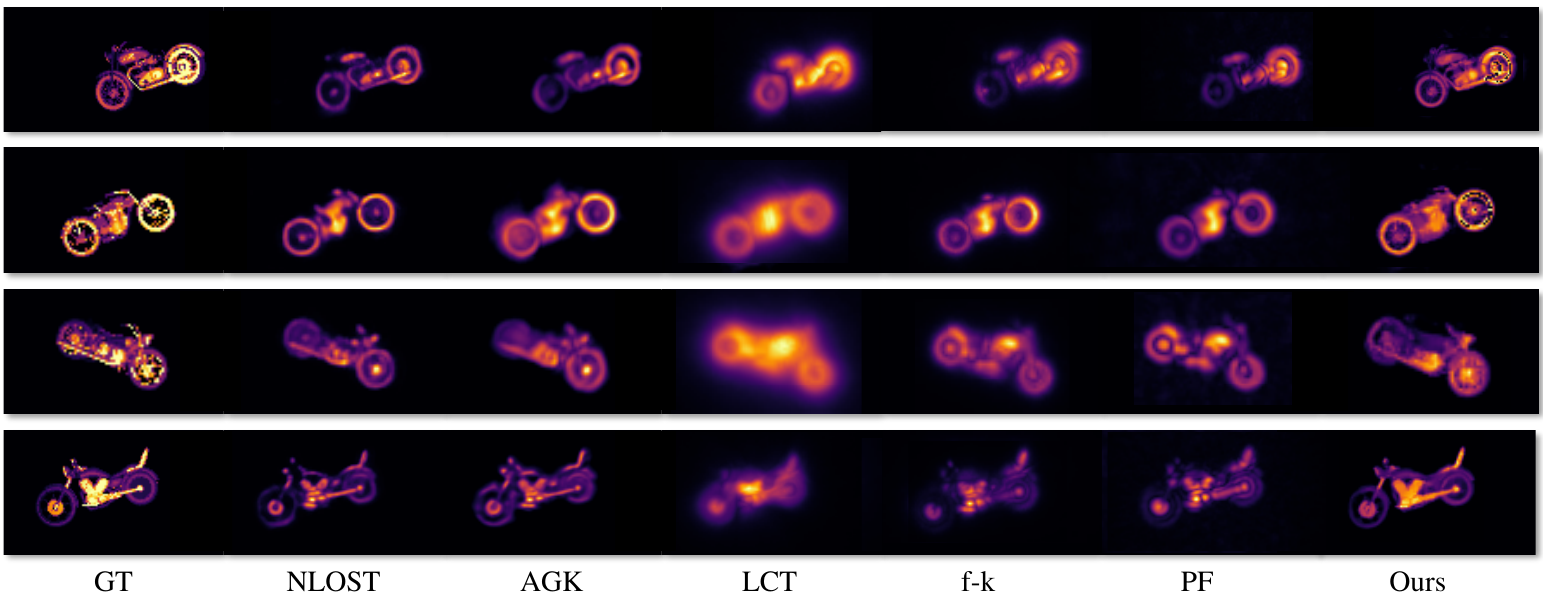}
\end{center}
\vspace{-13pt}
  \caption{\textbf{Qualitative comparison for albedo estimation task.} We compare MARMOT with LCT~\cite{otoole2018LCT}, f-k~\cite{lindell2019fk}, PF~\cite{liu2019PF}, NLOST~\cite{li2023nlost} and AGK~\cite{yu2023enhancing}. GTs are rendered using original meshes of motorbikes. MARMOT achieves reconstruction accuracy comparable to baseline methods. }
\label{fig:imaging2}
\vspace{-5pt}
\end{figure}

We also prvoide more comparison on downstream tasks, e.g. albedo estimation. 
We show the reconstruction results in Fig.~\ref{fig:imaging2}.

\begin{table}[t]
\centering
\begin{minipage}[t]{0.55\textwidth}
\centering
\small
\setlength{\tabcolsep}{3pt}
\label{tab:ablations}
\caption{Similarity comparison between original and recovered transients. We highlight the best in bold.} 

\resizebox{0.85\textwidth}{!}{
\begin{tabular}{cccccc} 
\toprule
\textbf{Parameter} & \textbf{Blocks/Dim.} & \textbf{ED$\downarrow$} & \textbf{CS$\uparrow$} & \textbf{SSIM$\uparrow$} & \textbf{PSNR$\uparrow$} \\
\midrule
\multirow{4}{*}{Encoder Depth} & 4 & 0.0038 & 0.9620 & 0.9704 & 36.31 \\
                               & 8 & 0.0038 & 0.9623 & 0.9707 & 36.33 \\
                               & 16 & \textbf{0.0036} & 0.9628 & \textbf{0.9711} & \textbf{36.40} \\
                               & 24 & 0.0037 & \textbf{0.9629} & 0.9705 & 36.39 \\
\midrule
\multirow{4}{*}{Encoder Width} & 128 & 0.0038 & 0.9618 & 0.9701 & 36.31 \\
                               & 256 & 0.0038 & 0.9616 & 0.9702 & 36.34 \\
                               & 512 & \textbf{0.0037} & \textbf{0.9631} & 0.9704 & \textbf{36.40} \\
                               & 1024 & \textbf{0.0037} & 0.9629 & \textbf{0.9705} & 36.39 \\

\midrule
\multirow{4}{*}{Decoder Depth} & 4 & 0.0053 & 0.9329 & 0.9637 & 35.08 \\
                               & 8 & \textbf{0.0037} & \textbf{0.9629} & \textbf{0.9705} & 36.39 \\
                               & 16 & 0.0043 & 0.9452 & 0.9667 & \textbf{36.41} \\
                               & 24 & 0.0044 & 0.9389 & 0.9601 & 35.97 \\
\midrule
\multirow{4}{*}{Decoder Width} & 64 & 0.0039 & 0.9618 & 0.9699 & 36.30 \\
                               & 128 & 0.0037 & 0.9627 & 0.9703 & 36.38 \\
                               & 256 & \textbf{0.0036} & 0.9628 & 0.9703 & \textbf{36.40} \\
                               & 512 & 0.0037 & \textbf{0.9629} & \textbf{0.9705} & 36.39 \\

\bottomrule
\end{tabular}}

\end{minipage}%
\hfill
\begin{minipage}[t]{0.32\textwidth}
\centering
\label{tab:structure}
\caption{Default parameters for MARMOT architecture.}

\resizebox{0.85\textwidth}{!}{
\begin{tabular}{@{}ll@{}}
\toprule
\textbf{Component} & \textbf{Parameters}\\ \midrule
Encoder Width & 1024 \\
Encoder Depth & 24 \\
Decoder Width & 512 \\
Decoder Depth & 8 \\
\# of Heads (Encoder) & 16 \\
\# of Heads (Decoder) & 16 \\
Mask Ratio & 0.95 \\
\midrule
Optimizer & AdamW \\
Learning Rate & 1e-4 \\
Beta Parameters & (0.9, 0.95) \\
\midrule
learning rate schedule & cosine decay \\
warm-up epochs & 6 \\
\bottomrule
\end{tabular}
}

\end{minipage}
\end{table}

\begin{table}[t]
\centering
\small
\setlength{\tabcolsep}{2pt}
\caption{\textbf{Quantitative comparison for Fig.~\ref{fig:comparesyth}.} Four evaluation metrics with best in bold.}
\label{tab:comparesyn2}
\resizebox{0.6\linewidth}{!}{
\begin{tabular}{ccccc}
\toprule
\textbf{Methods} & \textbf{ED$\downarrow$} & \textbf{CS$\uparrow$} & \textbf{SSIM$\uparrow$} & \textbf{PSNR$\uparrow$} \\
\midrule
LCT & 0.0858 & 0.8951 & 0.4790 & 17.48 \\
SOCR (BP) & 0.0841 & 0.7909 & 0.5312 & 16.00 \\
CC-SOCR (BP)& 0.0552 & 0.8762 & 0.7461 & 17.38 \\
MARMOT (LCT) & 0.0469 & \textbf{0.9934} & 0.6374 & 23.36 \\
MARMOT (f-k) & \textbf{0.0184} & 0.9796 & \textbf{0.7527} & \textbf{28.30} \\
MARMOT (PF) & 0.0419 & 0.9801 & 0.5461 & 23.97 \\
\bottomrule
\end{tabular}
}
\vspace{-10pt}
\end{table}

\begin{figure*}[t]
\begin{center}
\includegraphics[width=0.99\linewidth]{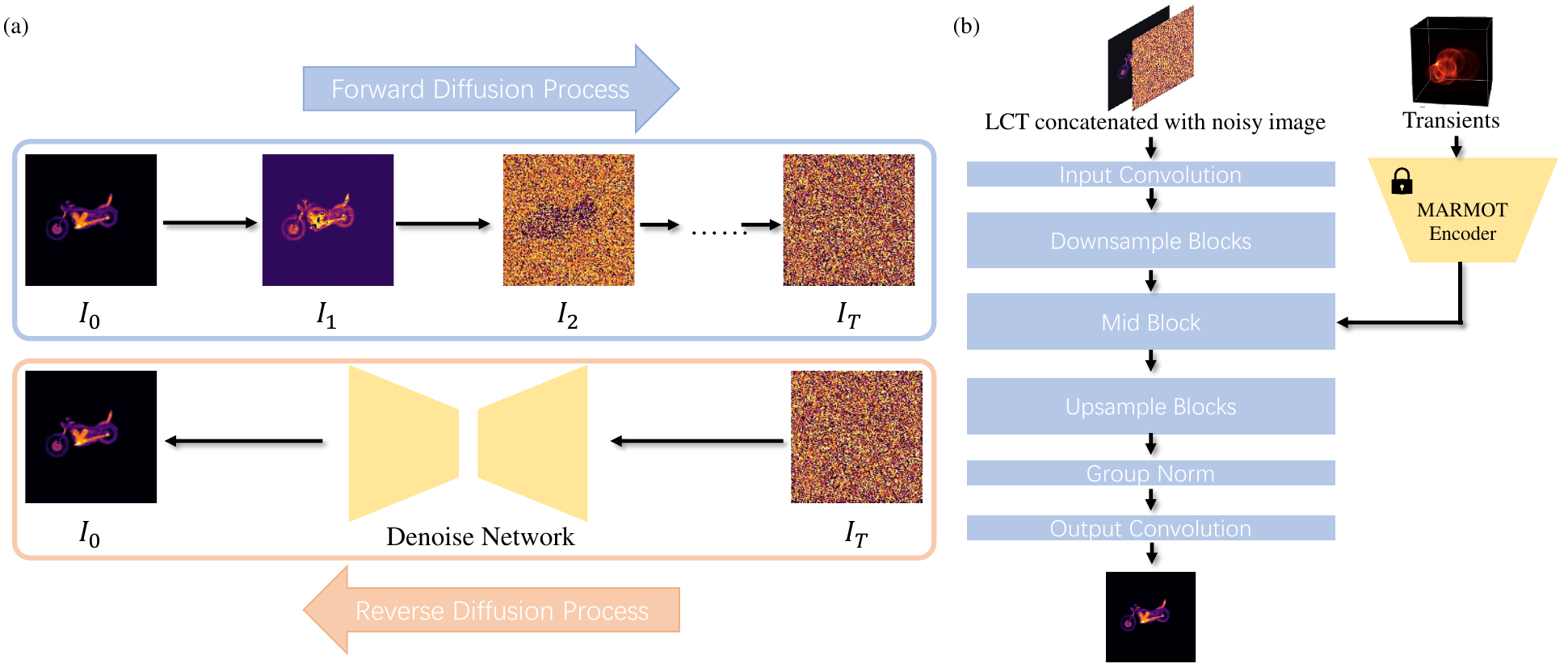}
\end{center}
  \caption{\textbf{Visualization of Diffusion Pipeline Architecture for Albedo and Depth Estimation. (a) shows the overwall diffusion process. (b) shows the architechture of the denoise network. Detailed parameters of the network are in Table.~\ref{tab:diffusion}}
}
\label{fig:downnet}
\end{figure*}

\section{Architecture for downstream tasks}
\label{sec:down1}
\subsection{Classification}
We use a simple neural network architecture for classification as previous work~\cite{chen2020learned,yu2023enhancing,li2023nlost}. We integrate a max pooling layer followed by a linear layer directly after the encoder of MARMOT. During training, we freeze the parameters in encoder and specifically optimize these two subsequent layers. 

\subsection{Albedo/Depth Estimation}
Albedo estimation and depth estimation can both be considered types of 2D inference tasks, which are among the most important and challenging tasks in non-line-of-sight (NLOS) imaging. Both tasks require the network to extract detailed information about hidden objects from transients. Early NLOS imaging networks utilized a U-Net architecture, which primarily relied on convolutional layers to extract and integrate features of various sizes for imaging purposes.

Most existing neural networks~\cite{chen2020learned,yu2023enhancing,li2023nlost} rely on a physically-based feature extraction module to provide a good initial state for the network and to guide and accelerate convergence. These modules leverage physical properties of the light transport in the scene to enhance the initial representations used by the network, facilitating more efficient learning.

The rapid advancement of diffusion models has recently provided new solutions for Non-Line-of-Sight (NLOS) imaging. Building upon prior work~\cite{su2024model} that combines diffusion models with NLOS imaging, we design our imaging pipeline. Specifically, during training, rather than directly training the network to predict the final image, our diffusion-based pipeline adopts a generative approach by using a denoising network to synthesize images from noise. The detailed process and results are illustrated in Fig.~\ref{fig:downnet}.

Our pipeline employs a lightweight U-Net architecture to predict and remove noise from the input. The network takes as input the concatenation of a noisy image and a coarse reconstruction obtained from LCT. Additionally, features extracted from transients by the MARMOT encoder are used as conditional information to guide the image generation process. We provide the specific parameters of each component in Table~\ref{tab:diffusion}.

\begin{table}[htbp]
\centering
\caption{Network Architecture of the denoise network.}
\vspace{5pt}
\begin{tabular}{l l l}
\toprule
\textbf{Layer Name} & \textbf{Layer Type} & \textbf{Details} \\
\midrule
Input convolution & Conv2d & channel:$2 \rightarrow 128$, kernel\_size=3, \\
                  &        & stride=1, padding=1 \\
Time Embedding & Linear layer & dimension 512 with SiLU activation \\
Downsample blocks & ResNet Blocks & 6 Downsample Blocks \\
\quad DownBlock2D (x2) & & ResnetBlock2D \\
\quad DownBlock2D (x2) & & ResnetBlock2D \\
\quad AttnDownBlock2D & & ResnetBlock2D With Attention \\
\quad DownBlock2D & & ResnetBlock2D \\
Mid block & ResNet Block & ResnetBlock2D with CrossAttn \\
Upsample blocks & ResNet Blocks & 6 Upsample Blocks \\
\quad UpBlock2D & & ResnetBlock2D \\
\quad AttnUpBlock2D & &ResnetBlock2D With Attention \\
\quad UpBlock2D (x4) & & ResnetBlock2D \\

GroupNorm & GroupNorm & num\_groups=32, num\_channels=128, \\
          &           & activation func=SiLU \\
Output convolution & Conv2d & channel:$128 \rightarrow 1$, kernel\_size=3, \\
          &           & stride=1, padding=1 \\
\bottomrule
\end{tabular}
\label{tab:diffusion}
\end{table}

\end{document}